\documentclass[letterpaper, 10 pt, journal, twoside]{ieeetran}  

\IEEEoverridecommandlockouts                              
\usepackage{graphics} 
\usepackage{epsfig} 
\usepackage{times} 
\usepackage{amsmath} 
\usepackage{amssymb}  
\usepackage{bbm}
\usepackage{mathtools}
\usepackage{multirow}
\usepackage{units}
\usepackage{bm}
\usepackage{xcolor}
\usepackage{hyperref}
\usepackage[ruled,vlined]{algorithm2e}
\SetCommentSty{mycommfont} 
\usepackage{cite}

\DeclareMathOperator{\diag}{diag}
\DeclareMathOperator{\motionupdate}{motion\_update}

\DeclareMathOperator{\mmentupdate}{mment\_update}

\newcommand{\SE}{\text{SE(3)}}
\newcommand{\SO}{\text{SO(3)}}
\newcommand{\se}{\mathfrak{se}(3)}

\newcommand{\bzero}{\mathbf{0}}
\newcommand{\beps}{\bm{\epsilon}}

\newcommand{\bSigma}{\bm{\Sigma}}
\newcommand{\bE}{\mathbf{E}}
\newcommand{\bO}{\mathbf{O}}
\newcommand{\bR}{\mathbf{R}}
\newcommand{\bT}{\mathbf{T}}
\newcommand{\bU}{\mathbf{U}}

\newcommand{\bp}{\mathbf{p}}

\newcommand{\bt}{\mathbf{t}}

\newcommand{\bz}{\mathbf{z}}

\newcommand{\wi}[1]{w^{(#1)}}
\newcommand{\bti}[1]{\bt^{(#1)}}
\newcommand{\bRi}[1]{\bR^{(#1)}}
\newcommand{\bTi}[1]{\bT^{(#1)}}

\newcommand{\cM}{\mathcal{M}}
\newcommand{\cN}{\mathcal{N}}

\newcommand{\cW}{\mathcal{W}}
\newcommand{\cX}{\mathcal{X}}

\newcommand{\bbR}{\mathbb{R}}

\DeclareMathOperator*{\argmin}{arg\,min}
\DeclareMathOperator*{\argmax}{arg\,max}
\let\diag\relax
\DeclareMathOperator*{\diag}{diag}

\title{
Probabilistic Visual Place Recognition for Hierarchical Localization
}

\author{Ming Xu, Niko Sünderhauf and Michael Milford
\thanks{Manuscript received: July, 24, 2020; Revised October, 8, 2020; Accepted November, 5, 2020.}
\thanks{This paper was recommended for publication by Editor Sven Behnke upon evaluation of the Associate Editor and Reviewers' comments. 
This work was supported by the QUT Centre for Robotics and ARC grant CE140100016.} 
\thanks{The authors are with the QUT Centre for Robotics, Queensland University of Technology, Brisbane, QLD 4000, Australia
        {\tt\footnotesize mingda.xu@hdr.qut.edu.au}}%
\thanks{© 2021 IEEE.  Personal use of this material is permitted.  Permission from IEEE must be obtained for all other uses, in any current or future media, including reprinting/republishing this material for advertising or promotional purposes, creating new collective works, for resale or redistribution to servers or lists, or reuse of any copyrighted component of this work in other works.}
\thanks{Digital Object Identifier (DOI): see top of this page.}
}

\newcommand{\githublink}{\url{https://github.com/mingu6/ProbFiltersVPR.git}}

\begin{document}

\maketitle
\markboth{IEEE Robotics and Automation Letters. Preprint Version. Accepted November, 2020}
{Xu \MakeLowercase{\textit{et al.}}: Probabilistic Visual Place Recognition for Hierarchical Localization}

\begin{abstract}

Visual localization techniques often comprise a hierarchical localization pipeline, with a visual place recognition module used as a coarse localizer to initialize a pose refinement stage. While improving the pose refinement step has been the focus of much recent research, most work on the coarse localization stage has focused on improvements like increased invariance to appearance change, without improving what can be loose error tolerances. In this letter, we propose two methods which adapt image retrieval techniques used for visual place recognition to the Bayesian state estimation formulation for localization. We demonstrate significant improvements to the localization accuracy of the coarse localization stage using our methods, whilst retaining state-of-the-art performance under severe appearance change. Using extensive experimentation on the Oxford RobotCar dataset, results show that our approach outperforms comparable state-of-the-art methods in terms of precision-recall performance for localizing image sequences. In addition, our proposed methods provides the flexibility to contextually scale localization latency in order to achieve these improvements. The improved initial localization estimate opens up the possibility of both improved overall localization performance and modified pose refinement techniques that leverage this improved spatial prior.



\end{abstract}


\section{Introduction}

\IEEEPARstart{T}{his} letter addresses the long-term visual localization problem, where the aim is to localize a camera within a pre-built map across varying levels of appearance change induced by time of day, weather, seasonal or structural changes.
Recent work around 6DoF visual localization (VLoc) \cite{Sattler2018} takes steps towards designing localization pipelines that are robust to appearance change for indirect, feature-based visual SLAM (VSLAM). One of the objectives in VLoc is to yield a precise 6DoF pose for a query image within a large-scale, point cloud reconstruction of a scene built from a set of reference images. This query image is captured under different appearance conditions to the reference images. \textit{Hierarchical} pipelines are commonly used \cite{Sarlin19, hloc2018, germain2019sparsetodense} where first, a Visual Place Recognition (VPR) \cite{lowry2016} system is used as a \textit{coarse localizer} to retrieve relevant reference images and subsequently a \textit{pose refinement} step is applied to the query image and the subset of the point cloud corresponding to the retrieved images to estimate the 6DoF pose. 

Using this initial VPR retrieval step has been shown to both improve the scalability of localization in large environments and furthermore, reduces the occurrence of incorrect data associations between points in the query image and the map \cite{Sarlin19, hloc2018, germain2019sparsetodense}. One of the key factors determining performance is how close the localization estimate provided by the initial retrieval step is to the true query pose, motivating the development of more \textit{accurate} place recognition techniques. This observation doubly holds for pose refinement steps based on direct image alignment, where aligning query and reference images with large baseline shifts \cite{Engel14} can cause convergence to a suboptimal solution.

Despite many significant advances in recent years around improving the robustness of the pose refinement step \cite{DeTone18, Stumberg20, Dusmanu2019CVPR}, little research has addressed improving the coarse localizer in this hierarchical context. While there is an extensive literature around improving the invariance of place recognition systems to appearance change \cite{SeqSLAM, lowry2016, Garg18, Hausler19, vysotska16}, these systems are not designed to localize at tight error tolerances (e.g. $\leq 5$m) more suitable for hierarchical localization. The research described in this letter sets out to address these current capability gaps.

\begin{figure}[t]
    \vspace{2.1mm}
    \centering
    \includegraphics[width=0.48\textwidth]{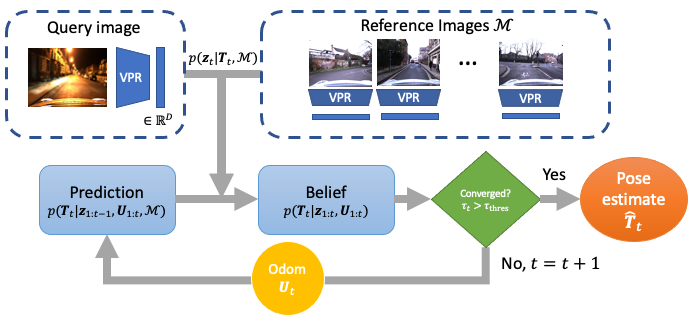}
    \vspace{-2mm}
    \caption{Our proposed methods recursively update the posterior state estimate of the robot given query images $\bz_t$ and odometry $\bU_{t-1}$ in the case of our MCL algorithm. At each time step, we check for convergence of the posterior and provide a pose estimate if it has converged.}
    \label{fig:process}
    \vspace{-5mm}
\end{figure}

Our contributions are summarized below:

\begin{enumerate}
    \item We propose two novel approaches to large-scale global localization over appearance change which involves adapting VPR methods based on image matching to the Bayesian state estimation framework. For each approach, this is achieved by specifying a suitable \textit{motion model} and \textit{measurement model}. The first method is based on a discrete Bayes' filter and localizes using appearance only to a topological map. The second method is a Monte Carlo Localization (MCL) algorithm which localizes using appearance and odometry. We show in our experiments that the MCL algorithm yields substantially better performance compared to the Bayes' filter at the cost of computation time.
    \item For each approach, we introduce a convergence detection and pose estimation method which is applied to the posterior state estimate at each time step. This allows our proposed methods to automatically scale the required localization latency (sequence length) based on the level of perceptual aliasing in the environment for a given level of precision.
    \item We demonstrate that our proposed localization algorithms can successfully localize sequences of query images with high recall and precision at \textit{significantly tighter error tolerances} across severe appearance change compared to existing methods based on template matching. This allows the possibility for our method to be used without following a full retrieval-refinement pipeline.
\end{enumerate}

We also make our code freely available online\footnote{\githublink}.

\section{Related works}

This section reviews relevant research on visual place recognition, Bayesian state estimation for localization in robotics and finally, hierarchical visual localization.

\subsection{Visual Place Recognition}

The VPR problem is commonly treated as a template matching problem; we assume the map consists of a set of ``places", and each place has an associated template extracted from either single images \cite{cummins08, Arandjelovic16, Torii15, Sunderhauf15} or image sequences \cite{SeqSLAM, Hausler19, vysotska16, maddern12catgraph, Doan2019ICCV, doan2019visual, facil2019condition}. Existing \mbox{VSLAM} systems already incorporate viewpoint-invariant VPR systems for loop-closure and localization \cite{Engel14, MurArtal15} and extending these systems to be robust to appearance change will help enable map reuse over extended periods of operation.

One focus within VPR research is around designing methods that can successfully match places across changing appearance conditions such as day/night and seasonal changes \cite{Torii15, Sunderhauf15, Arandjelovic16, olid2018single}, as well as across viewpoint changes \cite{galvezdbow2012, cummins08, Garg18}. In addition, these approaches can be further partitioned into methods that perform matching using deep learning \cite{Sunderhauf15, Arandjelovic16, olid2018single, RARS19} or handcrafted visual features \cite{Torii15, triggs05, arandjelovic13}. Common across many of these approaches is a match scoring system; given a pair of images, a typical VPR system outputs a ``quality score" which indicates the likelihood of two images being captured from the same place. Methods such as \cite{Arandjelovic16, Torii15, RARS19} first embed the images into an embedding space and set the quality score as the Euclidean distance between the embeddings; lower indicates a higher likelihood of a match. 

Our proposed approach treats the quality scores from a place recognition algorithm as noisy sensor measurements and localizes using the Bayesian state estimation framework. In principle, place recognition methods using alternative sensor modalities such as LiDAR can also be used with our proposed methods \cite{suaftescu2020kidnapped, gadd20, kimkim18, kim20}.

\subsection{Bayesian state estimation-based localization}

The Bayesian state estimation framework has been used extensively for localization in the past, however much of the initial work in this area involved measurement models designed for range-bearing sensors such as laser scanners \cite{Dellaert-1999, THRUN1999, Thrun2005}. More recently, \cite{brubaker16} used this framework to localize an autonomous vehicle using visual odometry in a large-scale cartographic map created using street segments. Furthermore, \cite{clark16} extends Monte Carlo Localization to multiple sensor modalities including vision. 

In the context of VPR, FAB-MAP \cite{cummins08} formulates topological mapping and localization as a Bayesian state estimation problem and uses handcrafted visual features to represent a place. This base model has been extended to include odometry and image sequences \cite{maddern12catslam, maddern12catgraph}. In comparison, our approach does not perform mapping, however unlike \cite{cummins08, maddern12catslam, maddern12catgraph}, it can successfully localize in a large-scale map across severe appearance change when used in conjunction with a condition invariant VPR method. Finally, \cite{Doan2019ICCV, doan2019visual} also uses a Bayesian state estimation framework for localization using VPR given by \cite{Torii15} however they do not test their method across severe appearance change for real-world datasets.

\subsection{Hierarchical visual localization}

An increasing body of research addresses building hierarchical localization pipelines \cite{Sarlin19, hloc2018, germain2019sparsetodense} for sparse, indirect feature-based maps, where VPR techniques form the initial coarse localizer. The pipeline typically involves an image retrieval step where the top-$k$ most similar reference images for a given query are retrieved. The pose refinement step is then performed on the subset of 3D points corresponding to the retrieved images. This methodology imposes a bottleneck on localization performance caused by the VPR system because a failed retrieval step provides a poor initialization for the pose refinement method.

Despite recent advances around the pose refinement stage \cite{DeTone18, Stumberg20, Dusmanu2019CVPR}, to the best of our knowledge there is minimal research on improving the VPR back-end's achievable error tolerances in this hierarchical context. We fill this gap with our proposed sequence-based back-ends and demonstrate high precision-recall performance at error tolerances suitable for initializing a pose refinement technique. Finally, we note that there is limited literature in the VLoc literature (as opposed to topological mapping) around localizing image sequences in general. \cite{Sattler2018} showed substantial performance gains using image sequences and/or multiple camera views in the pose refinement stage and \cite{clark17} explores using image sequences in the absolute pose regression context.  

\section{Bayesian state estimation for localization}

\subsection{Notation} 

A camera pose is given by a rigid-body transformation $\bT\in \SE$ between the world reference frame and the camera, represented by the $4\times4$ homogeneous matrix
\begin{equation*}
    \bT = \begin{pmatrix}
        \bR & \bt \\
        \mathbf{0} & 1
        \end{pmatrix},
\end{equation*}
where $\bR\in \SO$ and $\bt\in\bbR^3$. We represent linearized pose increments as Lie algebra elements $\beps \in \se$ which are in turn represented directly as vectors $\beps\in\bbR^6$ for convenience. We define the usual composition operator
\begin{equation*}
    \circ : \SE \times \se \rightarrow \SE, \quad \bT \circ \beps = \bT \cdot \exp \hat{\beps}.
\end{equation*}

\subsection{Overview}\label{ssec:overview}

In this section, we provide an overview around using Bayesian state estimation for localization. Suppose we have a sequence of noisy sensor measurements given by a query image sequence $\{\bz_s\}_{s=1}^t$ generated from corresponding unknown poses $\{\bT_s\}_{s=1}^t$, potentially with odometry estimates $\{\bU_s\}_{s=1}^{t-1}$ between measurements (e.g. from stereo visual odometry). For the purpose of this letter, we let $\bT_s, \bU_s\in \SE$, and $\bz_s\in\bbR^D$ be the query image embedding returned by a VPR system (e.g. NetVLAD \cite{Arandjelovic16}). Also, let the pre-built map of the environment be given by $\cM=\{(\bT_s^r, \bz_s^r)\}_{s=1}^N$ which consists of reference image embeddings $\bz_s^r$ and corresponding poses $\bT_s^r$. Bayesian state estimation seeks to estimate the \textit{posterior} distribution over the current pose $\bT_t$ given the map and the full sequence of sensor measurements and odometry through the recursion
\begin{align}\label{eq:bayesrecursion}
    p(\bT_t | \bz_{1:t}, \bU_{1:t}, \cM) \propto p(&\bz_t|\bT_t, \cM)p(\bT_t | \bT_{t-1}, \bU_{t-1}) \nonumber\\
    &p(\bT_{t-1} | \bz_{1:t-1}, \bU_{1:t-1}, \cM).
\end{align}
The implementation of this recursion depends on the underlying \textit{motion} and \textit{measurement} models which are given by $p(\bT_t | \bT_{t-1}, \bU_{t-1})$ and $p(\bz_t|\bT_t, \cM)$, respectively. The motion model encodes the dynamics of the robot and can be conditioned on an odometry estimate while the measurement model scores the likelihood of an observation given a robot pose. In this letter, we propose \textbf{two} algorithms using different model formulations: A Topological Localizer based off of a discrete Bayes filter which localizes using appearance information only (Section \ref{sec:topological}), and a Monte Carlo Localization (MCL) algorithm which utilizes appearance and odometry during localization (Section \ref{sec:mclocalization}). Figure \ref{fig:process} illustrates the full localization pipeline for our proposed methods.

\begin{figure}[t]
    \centering
    \includegraphics[width=0.48\textwidth]{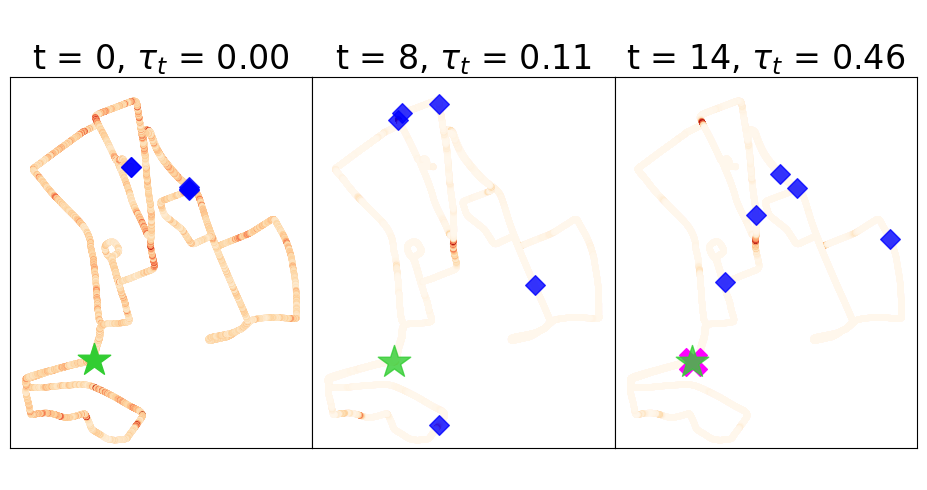}
    \includegraphics[width=0.48\textwidth]{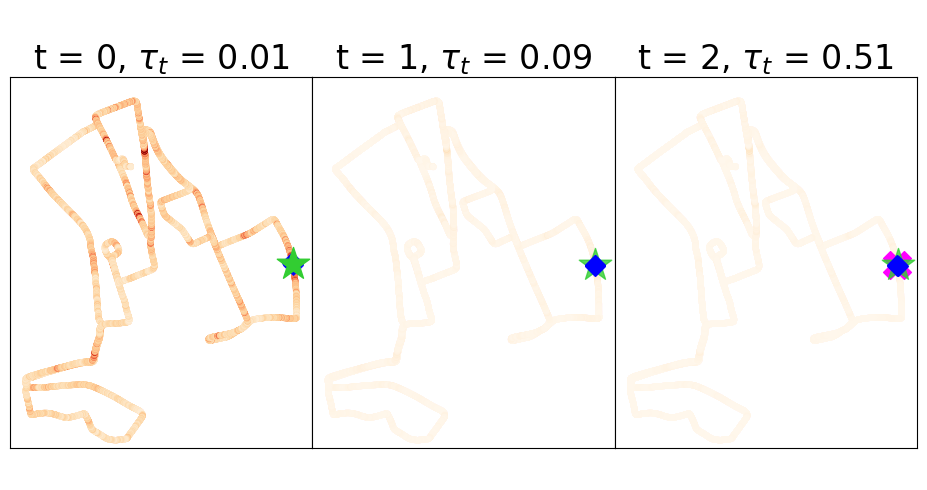}
    \vspace{-5mm}
    \caption{Demonstration of our proposed methods with our topological localizer; green represents the ground truth pose, blue represents the top 5 image retrieval candidates from VPR (NetVLAD) and magenta represents the model proposal. Darker shades of red indicate a higher posterior belief. \textbf{Top:} Query sequence with perceptual aliasing where VPR yields inconsistent retrievals for the first 10 timesteps, causing the posterior to not converge and consequently not localize (left, middle). After more iterations, more consistent retrieval candidates concentrates the posterior in the correct region in the map, yielding a correct pose estimate upon localization (right). \textbf{Bottom:} In this case, VPR works extremely well and our method converges to the correct pose estimate almost immediately.}
    \label{fig:confidence}
    \vspace{-5mm}
\end{figure}

\subsection{Convergence detection and pose estimation}

For each of our two proposed algorithms, we estimate the full posterior distribution over the robot pose. A desirable capability of a localization system is to yield a \textit{single pose estimate} from the posterior, which can then be used either as a final pose estimate or to initialize a pose refinement technique. We introduce a heuristic to detect convergence of the posterior to a single mode for both methods, which in practice yields extremely accurate localization performance, even before resorting to pose refinement. Figure \ref{fig:confidence} demonstrates our system running where VPR fails as well as when VPR works well, showing how our system scales localization latency to achieve state-of-the-art localization performance.


\section{Topological localizer}\label{sec:topological}

In this section, we introduce our first proposed method, the \textit{topological localizer}. We will first introduce the model formulation which is based off of a discrete Bayes filter and subsequently introduce the full localization algorithm. 

\subsection{Model formulation}

In our topological localizer, we assume the robot can only assume a finite set of poses which are given by the reference poses, formally $\bT_t \in \{\bT_s^r\}_{s=1}^N$. Consequently, our posterior at time $t$ is a probability vector $\bp_t$. In addition, we assume the reference images are captured in sequence; reference poses captured at nearby indices correspond to nearby poses. From this, we can define our motion model using discrete state transitions and use the forward algorithm for Hidden Markov Models to recursively update the posterior.

\subsection{Motion model}

We use a simple motion model for our topological localizer which does not require odometry estimates between query images. We assume the robot can transition to nearby poses corresponding to nearby frames in the video sequence with uniform probability at each time step. Formally,
\begin{equation}\label{eq:hmmtransition}
    p(\bT_{i}^r | \bT_{j}^r) \propto 
    \begin{cases}
    1 & w_l \leq i - j \leq w_u \\
    0 & \text{otherwise}
    \end{cases},
\end{equation}
where $w_u$ and $w_l$ are the upper and lower bounds for state transitions. Having sufficiently wide bounds allows for variations in velocity between query and reference traverses.

\subsection{Measurement model} \label{ssec:topomeas}

We can score the likelihood of observing a query image embedding $\bz_t$ at one of the finite valid poses $\bT_s^r$ using the Euclidean distance between the reference and query embeddings. Formally, our measurement model is given by
\begin{equation}\label{eq:hmmobs}
    p(\bz_t | \bT_s^r, \cM) \propto g(\bz_t, s, \cM) = \exp(-\lambda_1 \|\bz_t - \bz_{s}^r\|_2),
\end{equation}
where $\lambda_1$ is set automatically upon initialization given a user-set parameter $\delta > 0$. $\delta$ sets the intensity of the effect of sensor measurements on the posterior. At $t=1$, we set 
\begin{equation*}\label{eq:calibrate}
    \lambda_1 = \frac{\log \delta}{d_{0.975} - d_{0.025}},
\end{equation*}
where $d_q$ is the $q^{th}$ quantile of the set of distances between the first query embedding $\bz_1$ and all reference embeddings. Setting $\lambda_1$ in this way ensures the likelihoods in \eqref{eq:hmmobs} are consistently scaled (in a relative sense) across embedding choice and query/reference conditions. In principle, alternative VPR methods that directly score image pairs without an intermediate embedding stage such as \cite{Garg18} can also be used in place of the Euclidean distance between embeddings. 

\subsection{Posterior update}

Let $\bE\in\bbR^{N\times N}$ be the transition matrix, where the $i^{th}$ row and $j^{th}$ column entry is given by $\bE_{ij} = p(\bT_j^r | \bT_i^r)$. Let $\bO_t\in\bbR^{N\times N}$ be a diagonal matrix where $\bO_{t, ii} = g(\bz_t, i, \cM)$. The recursive update on the posterior from \eqref{eq:bayesrecursion} is given by
\begin{equation*}\label{eq:HMMbeliefupdate}
    \bp_t \propto \bO_t \bE^\top \bp_{t-1}.
\end{equation*}
The banded diagonal structure of $\bE$ implied by \eqref{eq:hmmtransition} ensures that this operation is linear with $N$.

\subsection{Convergence detection and pose estimation}\label{ssec:topoconverge}

Finally, we can use the posterior distribution to determine convergence of the filter and subsequently yield a single pose estimate. Our convergence detection measure $\tau_t\in(0,1]$ is given by the total probability mass in a neighborhood around the maximum a-posteriori (MAP) state estimate. Let $s^* = \argmax_{s} p(\bT_s^r | \bz_{1:t}, \cM)$ and from this,
\begin{equation}\label{eq:hmmscore}
    \tau_t = \sum_{s \in \cW} p(\bT_s^r | \bz_{1:t}, \cM), \quad \cW = \{s^* - w, \dots, s^* + w\}.
\end{equation}
If $\tau_t > \tau_{thres}$, then our pose estimate is $\widehat{\bT}_t = \bT_{\hat{s}}^r$ where
\begin{equation*}\label{eq:hmmproposal}
    \hat{s} = \Biggl\lfloor \sum_{s\in \cW} s p(\bT_s^r | \bz_{1:t}, \cM)\Biggr\rfloor.
\end{equation*}
We found empirically that averaging over a neighborhood of the MAP performed better than using the MAP directly.

\section{6DoF Monte Carlo Localization} \label{sec:mclocalization}

In this section, we introduce our second proposed method, the \textit{Monte Carlo Localization} (MCL) algorithm. In addition to appearance, it utilizes odometry estimates between query images to localize and in our experiments, demonstrates improved localization performance at the cost of computation time. We will first introduce the model formulation, followed by the posterior update steps. Next, we will outline the model components required to perform the posterior update and finally the convergence detection and pose estimation method given the estimated posterior.

\subsection{Model formulation} \label{ssec:mclformulation}

As opposed to the topological localizer, our MCL algorithm assumes the robot pose is continuous and can span all of $\SE$. Particle filters estimate the posterior distribution over the robot pose $p(\bT_{t}  |\bz_{1:t}, \bU_{1:t-1}, \cM)$ using a finite set of \textit{weighted samples} $\cX_t = \{\wi{i}_t, \bTi{i}_t\}_{i=1}^M$.

\subsection{Posterior update}

After particle initialization, the recursion given in \eqref{eq:bayesrecursion} is now implemented by first updating the particles by sampling from the motion model (motion update) and subsequently reweighting the particles using the measurement model (measurement update). To prevent sample degeneracy, we use a heuristic where systematic resampling \cite{doucet2009tutorial} is applied to particles after the update steps if the Effective Sample Size (ESS) given by $\text{ESS} = 1/\sum_{i=1}^M {\wi{i}_t}^2$ is below a threshold $p\in(0,1)$. We use $p=0.3$ in our work.

\subsection{Motion model}\label{ssec:motionupdate}

We apply a motion update to each particle given by
\begin{equation*}\label{eq:motionupdate}
    \motionupdate(\bU_{t-1}, \bTi{i}_{t-1}) = \bU_{t-1} \cdot \bTi{i}_{t-1} \circ \beps,
\end{equation*}
where $\beps \sim \mathcal{N}(\bzero, \bSigma_{odom})$. This is a simple, platform agnostic motion model that perturbs the screw motion given by the raw odometry reading with independent Gaussian noise.

\subsection{Measurement model}\label{ssec:mmentupdate}

The measurement model updates a particle weight $\wi{i}_{t-1}$ using the likelihood of observing the current query descriptor $\bz_t$ at the pose $\bTi{i}_t$. Our measurement update incorporates VPR measurements similarly to \eqref{eq:hmmobs}, but also includes the physical distance between the particle and nearby reference images. From this, we introduce our measurement update as
\begin{align}\label{eq:sensorupdateapprox}
    \mmentupdate(\bz_t, \bTi{i}_t; \cM) = &\sum_{k=1}^K \exp\bigl(- \lambda_1 \|\bz_t - \bz_{n_k}^r\|_2  \nonumber \\
    & -\lambda_2 d(\bTi{i}_t, \bT_{n_k}^r) \bigr),
\end{align}
where $\lambda_1$ is as per \eqref{eq:hmmobs} and $\lambda_2 > 0$ is set by the user and controls the impact of map deviation on the particle weight. Furthermore, $d$ is a metric on $\SE$ given by
\begin{equation}\label{eq:se3metric}
    d(\bT_1, \bT_2) = \|\bt_1 - \bt_2\|_2 + \alpha \|\log(\bR_1^\top \bR_2) \|_2,
\end{equation}
where $\|\log(\bR_1^\top \bR_2) \|_2$ is the minimum rotation angle between $\bR_1$ and $\bR_2$ and $\alpha$ is a scaling weight. $K$ is the number of nearest neighbor map reference images to retrieve for each particle and $n_{k}$ is the index of the $k^{th}$ nearest reference image according to \eqref{eq:se3metric} to the particle pose. For \eqref{eq:se3metric}, fast nearest-neighbor search algorithms are available \cite{Ichnowski14}.

\subsection{Particle Initialization}\label{ssec:initializeparticles}
We initialize our $M$ particles from the set of reference image poses $\{\bT_{i}^r\}_{i=1}^N$ with probability proportional to the visual similarity from $\bz_1$ and perturb the poses with Gaussian noise. Formally, for $\beps \sim \cN(\bzero, \bSigma_\text{init})$, let
\begin{equation*}
    \bTi{i}_1 = \bT_{n}^r \circ \beps , \quad n \sim \text{Cat}(N, \bp) \quad
\end{equation*}
where $\text{Cat}(N, \bp)$ is the categorical distribution over the integers from 1 to $N$ with probability vector $\bp$ where $\bp_i = \eta \exp(-\lambda_1 \|\bz_t - \bz_{i}^r\|_2)$ and $\eta$ is a normalizing constant. 

\subsection{Convergence detection and pose estimation}\label{ssec:mclconverge}

To determine convergence of the filter from the particle set $\widehat{\bT}_t$, we follow the same process as in Section \ref{ssec:topoconverge} where we compute the total probability mass in a neighborhood around the MAP state estimate. Let $i^* = \argmax_{i} \wi{i}_t$, $\cW = \{i: d(\bTi{i}_t, \bTi{i^*}_t) < r\}$ and $\tau_t = \sum_{i\in\cW} \wi{i}_t$. If $\tau_t  > \tau_{thres}$, then we assume the filter has converged and estimate the final pose $\widehat{\bT}_t = \begin{pmatrix}
        \widehat{\bR} & \hat{\bt} \\
        \mathbf{0} & 1
        \end{pmatrix}$, where
\begin{equation*}\label{eq:mclproposal}
    \hat{\bt} = \frac{1}{\tau_t}\sum_{i\in\cW} \wi{i}_t\bti{i}_t, \quad \widehat{\bR}_t = \argmin_{\bR} \sum_{i\in\cW} \|\bR - \frac{\wi{i}_t}{\tau_t} \bRi{i}_t\|_F.
\end{equation*}

\section{Experimental setup}

This section details the experimental setup and localization performance of our proposed methods.

\subsection{Datasets}
 
\begin{figure}[t]
    \centering
    \includegraphics[width=0.48\textwidth]{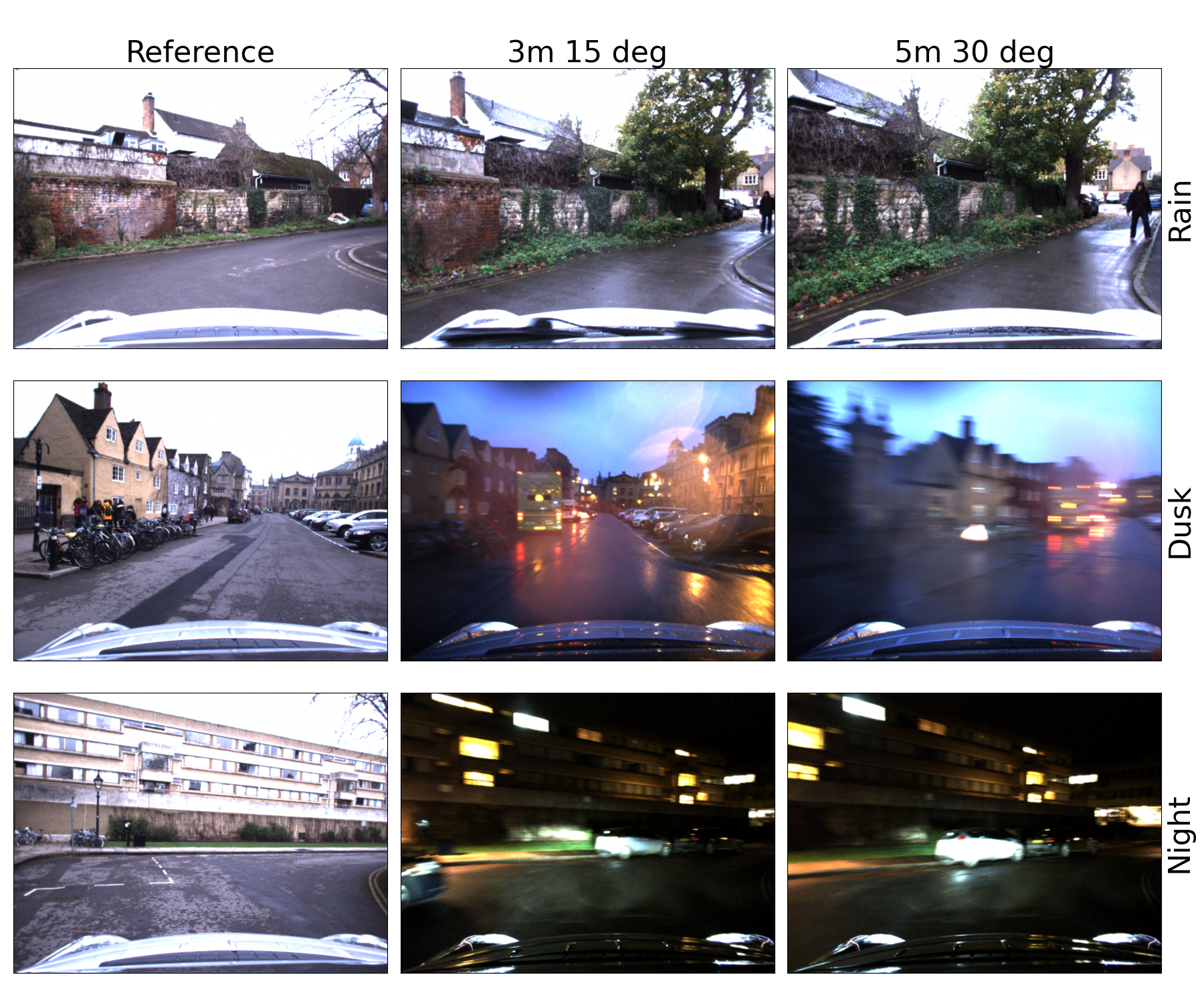}
    \vspace{-3mm}
    \caption{Sample images from the RobotCar dataset illustrating levels of condition change and baseline shifts for selected error tolerances. \textbf{Left:} Reference images. \textbf{Middle/Right:} Images from query traverses furthest from reference image that are still within the given metric error tolerance.}
    \vspace{-5mm}
    \label{fig:sampleimages}
\end{figure}

We use the Oxford RobotCar dataset \cite{RobotCarDatasetIJRR}, which is comprised of traverses of Oxford city during different seasons, time of day and weather conditions. We use the forward facing camera for four overlapping traverses spaced months apart over a wide range of appearance conditions; three for query and one for reference, each approximately 10km in length. The pose estimates for images in the reference traverse are given by the RTK GPS \cite{RobotCarRTKarXiv} and subsampled at approximately 0.5m increments yielding 13,595 images. We performed 500 trials where for each trial we localize a short query sequence with a randomly selected starting location within the full query traverse. Each short query sequence contains 30 images spaced approximately 3m apart. We use the stereo visual odometry (VO) provided with the RobotCar dataset as odometry. Finally, we use the implementation of NetVLAD \cite{Arandjelovic16} found in \cite{Cieslewski18icra} and DenseVLAD \cite{Torii15} as image embeddings and extracted these from full resolution images. Table \ref{tab:datasets} summarizes the datasets used and Figure \ref{fig:sampleimages} shows example images in the RobotCar dataset.

\begin{table}[ht]
\caption{Traverses used in the experiments}
\vspace{-3mm}
\label{tab:datasets}
\begin{center}
\begin{tabular}{|c||c|c|}
\hline
\textit{Name} &  \textit{Conditions} & \textit{Traverse} \\
\hline
\textit{2015-03-17-11-08-44} & Overcast & Reference\\
\hline
\textit{2015-10-29-12-18-17} & Rain & Query \\
\hline
\textit{2014-11-21-16-07-03} & Dusk, Rain & Query \\
\hline
\textit{2014-12-16-18-44-24} & Night & Query\\
\hline
\end{tabular}
\end{center}
\vspace{-5mm}
\end{table}

\subsection{Model Comparisons}

For our experiments, we use precision-recall (PR) curves to compare our proposed methods with state-of-the-art sequence-matching methods \cite{SeqSLAM, vysotska16} as well as a single image matching baseline. Parameter tuning was performed and modifications were applied to all comparison methods to maximize their competitiveness with our proposed approach to make the comparison as relevant as possible. 

Precision and recall are computed across all 500 trials. For each trial, we note whether or not a method successfully localizes (e.g. the convergence criteria is met for our methods). A successfully localized trial is defined to be a \textit{true positive} if the ground truth error falls within a specified error tolerance and a \textit{false positive} if the error falls outside of the tolerance. A \textit{false negative} is recorded if localization failed. In this work, we use two error tolerances; one with 3 meters translation and 15 degrees orientation error more suitable for initializing direct image alignment and the other with 5 meters translation and 30 degrees orientation error more suitable for indirect, feature-based methods. Both tolerances ensure that query and matched reference images have high covisibility necessary for pose refinement, maximizing the relevance of the results to hierarchical localization methods. We define the translation error as $\|t_p - t_{gt}\|_2$ and orientation error for each trial to be $\|\log(\bR_p^\top \bR_{gt})\|$ when measuring error between proposals and the ground truth pose.

\subsubsection{Proposed Approaches} For our proposed approaches if $\tau_t>\tau_\text{thres}$ at time $t$, then we evaluate the error between the RTK query image pose and the pose estimate provided by our methods. PR curves are generated by varying $\tau_\text{thres}$.

\subsubsection{Single Image Matching}

Our single image matching baseline localizes a trial by retrieving the reference image closest in appearance to the first query image in the sequence. Ground truth error is calculated using the RTK poses of the two images. Precision-recall (PR) curves are generated by varying a non-matching threshold; matches are accepted only if the embedding distance is lower than this threshold.

\subsubsection{Sequence Matching}

We compare against a customized version (to maximize performance) of the sequence matching method in \cite{SeqSLAM} and compute an image-to-image distance matrix between a query sequence and reference traverse from our image embeddings. A linear search for coherent sequence matches is performed within this matrix to allow for changes in velocity between and within traverses. Ground truth error is calculated between the pose of the first images of both the query and the successfully matched reference sequence. PR curves are generated by varying a quality score; see \cite{SeqSLAM} for details. Furthermore, a fixed sequence length of 10 was found to maximize performance across all traverses. Contrast normalization of the distance matrix used in \cite{SeqSLAM} was not applied, resulting in further significant performance gains.

\subsubsection{Graph-based Matching}

We also compare against the online, graph-based sequence matching approach from \cite{vysotska16}, which removes the constraint of the linear search of the previous sequence matching method. In addition, the online nature of their method allows a more direct comparison to our methods compared to fixed-length sequence matching. We found that the expansion rate parameter $\alpha=0.98$ yielded an optimal balance between computation time and performance across all traverses.

At each time step, the graph-based approach yields the best match within the reference images for the current query image in the sequence. A non-matching score $\hat{w}$ similar to the single image matching case is used to accept/reject this match. Ground truth error is calculated between the pose of the query and reference images corresponding to the first accepted match. PR curves are generated by varying $\hat{w}$.

\subsubsection{MCL Ablations}

We test our MCL algorithm replacing odometry from VO with the RTK GPS ground truth to show how higher quality odometry estimates affect localization performance. In another test, we discarded the visual similarity term in \eqref{eq:sensorupdateapprox}, using proximity to reference images poses only but observed that localization consistently failed.

\subsection{Parameter Values}

For the topological filter, we used $\delta = 5$, $w_l = -2$, $w_u = 10$ and the window parameter in \eqref{eq:hmmscore} set to $w=6$. We list the parameter values used for our MCL algorithm in Table \ref{tab:parameters}. The same parameter values were used across all query/reference/embedding combinations. 

\begin{table}[ht]
\caption{MCL parameter values}
\vspace{-3mm}
\label{tab:parameters}
\begin{center}
\begin{tabular}{|c|c|c|}
\hline
\textit{Parameter} & \textit{Section} & \textit{Value} \\
\hline
$M$ & \ref{sec:mclocalization} & 6000 \\
\hline
$\delta$ & \ref{ssec:topomeas} & 5 \\
\hline
$\diag(\bSigma_{init})$ & \ref{ssec:motionupdate} & $[2.0^2, 0.5^2, 0.5^2, 0.05^2, 0.05^2, 0.1^2]$ \\
\hline
$\diag(\bSigma_{odom})$ & \ref{ssec:motionupdate} & $[0.8^2, 0.3^2, 0.3^2, 0.04^2, 0.04^2, 0.08^2]$ \\
\hline
$\lambda_2$ & \ref{ssec:mmentupdate} & 0.2 \\
\hline
$K$ & \ref{ssec:mmentupdate} & 3\\
\hline
$\alpha$ & \ref{ssec:mmentupdate} & 15 \\
\hline
$r$ & \ref{ssec:mclconverge} & 10 \\
\hline
\end{tabular}
\end{center}
\vspace{-6mm}
\end{table}

\section{Experimental results}

\subsection{Localization Performance}

\begin{figure*}[tb]
    \vspace{2.1mm}
    \centering
    \includegraphics[width=0.48\textwidth]{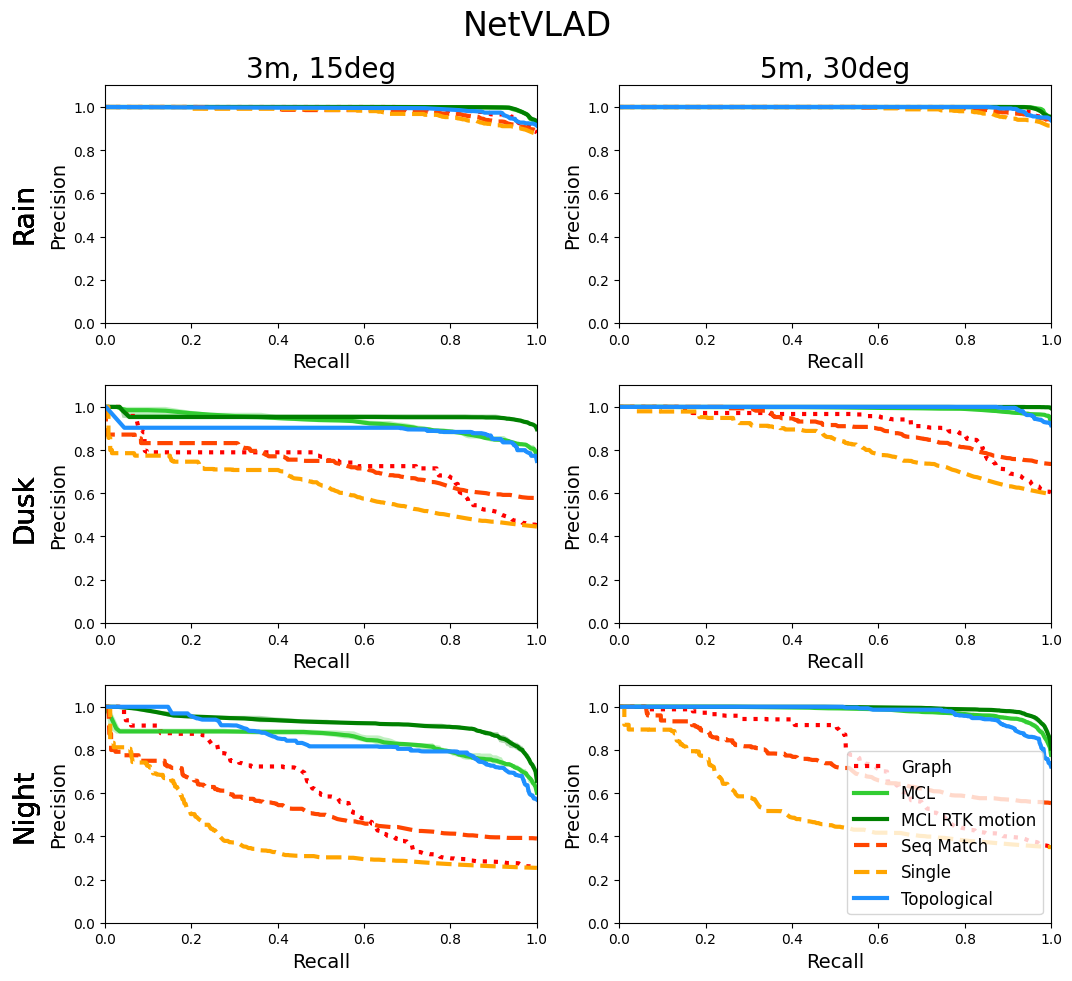}
    \includegraphics[width=0.48\textwidth]{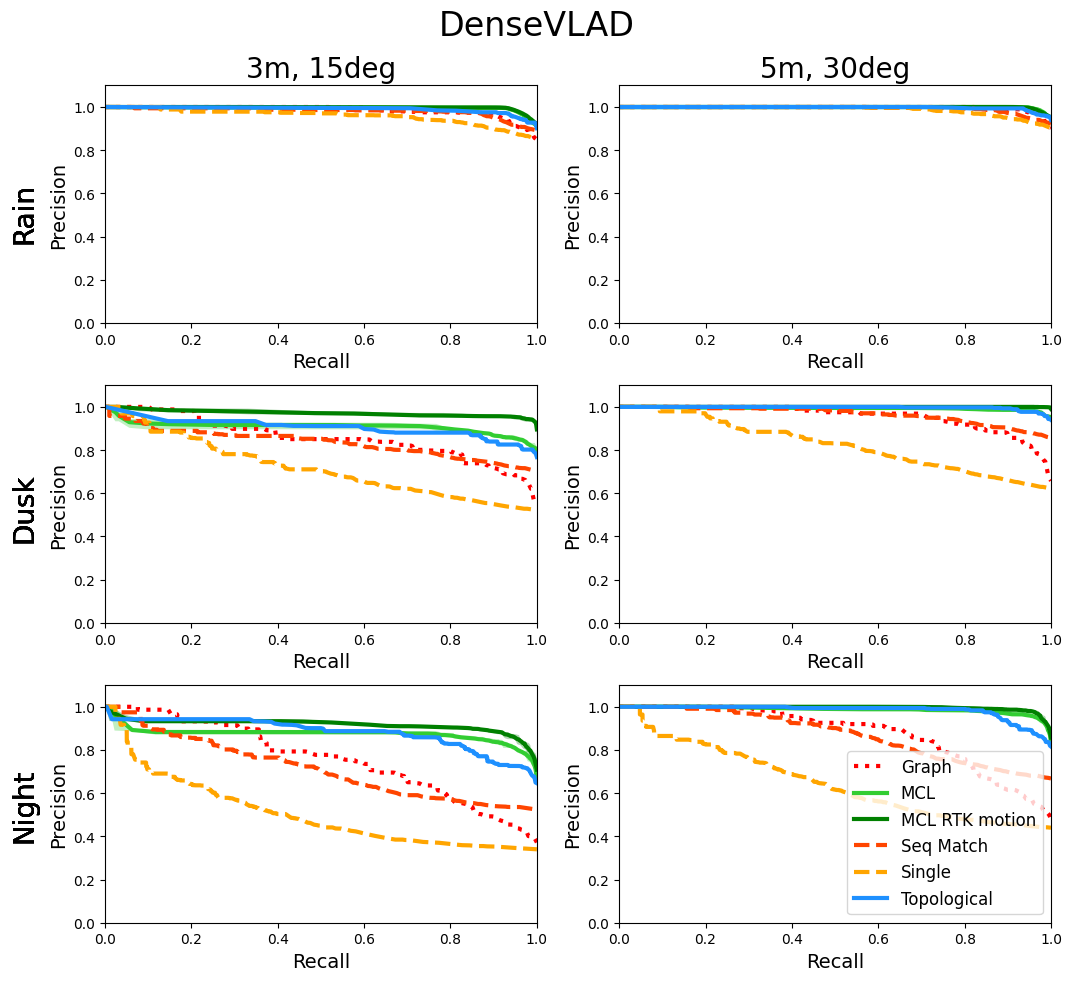}
    \vspace{-2mm}
    \caption{Precision-recall curves for all experiments. For the MCL-based algorithms, we shade in the region between the max and min precision values for each recall to account for variability for our MCL algorithm over 5 trials. The proposed topological filter and MCL methods substantially outperform all other methods in all cases. In addition, the inclusion of quality odometry estimates improved performance substantially, especially at the finer tolerance. In some cases, the topological filter outperformed the MCL algorithm with VO, suggesting some sensitivity of performance to the quality of the odometry.}
    \vspace{-4mm}
    \label{fig:PR}
\end{figure*}

We measure performance using PR curves because they capture the performance characteristics of a coarse localizer. We present the generated PR curves in Figure \ref{fig:PR}. Recall at 99\% precision and area under the curve metrics for all methods are also presented for a more direct numerical comparison in Tables \ref{tab:maxrecallNV}, \ref{tab:aucNV} and \ref{tab:aucDV}. 

Our proposed methods perform substantially better overall compared to the comparison methods, especially across severe appearance change. This is achieved in part due to our Bayesian state-estimation formulation because we choose to localize query sequences only when the posterior belief converges. For query sequences with higher levels of perceptual aliasing with inconsistent VPR retrievals, the posterior converges more slowly but as a consequence yields more accurate and reliable pose estimates. In contrast, linear or greedy sequence matching used in the comparison approaches tend to fail under these circumstances. 

\begin{table}[ht]
\caption{Recall (\%) at 99\% precision (5m, 30 deg)}
\vspace{-3mm}
\label{tab:maxrecallNV}
\begin{center}
\begin{tabular}{|c||c|c||c|c||c|c|}
\hline
\multirow{2}{*}{Model} & \multicolumn{2}{c||}{\textit{Rain}} & \multicolumn{2}{c||}{\textit{Dusk}} &\multicolumn{2}{c|}{\textit{Night}}\\
 \cline{2-7}
 & \textit{NV} & \textit{DV} & \textit{NV} & \textit{DV} & \textit{NV} & \textit{DV} \\\hline
\textit{Single} & 65.8 & 70.8 & 4.0 & 9.4 & 1.2 & 4.8 \\
\hline
\textit{SeqM} & 83.1 & 83.5 & 31.1 & 47.4 & 6.2 & 22.2 \\
\hline
\textit{Graph} & 88.3 & 89.7 & 14.0 & 42.6 & 6.0 & 30.7 \\
\hline
\textit{Topological} & 91.7 & 93.6 & 92.0 & 90.9 & 57.6 & 80.0 \\
\hline
\textit{MCL} & \textbf{97.2} & \textbf{96.6} & 80.0 & 80.2 & 55.8 & 47.3 \\
\hline
\textit{MCL (RTK)} & 96.2 & 96.0 & \textbf{100.0} & \textbf{100.0} & \textbf{76.1} & \textbf{86.4} \\
\hline
\end{tabular}
\end{center}
\vspace{-7mm}
\end{table}

\begin{table}[ht]
\caption{Area-under-curve (NetVLAD)}
\vspace{-3mm}
\label{tab:aucNV}
\begin{center}
\begin{tabular}{|c||c|c||c|c||c|c|}
\hline
\multirow{2}{*}{Model} & \multicolumn{2}{c||}{\textit{Rain}} & \multicolumn{2}{c||}{\textit{Dusk}} &\multicolumn{2}{c|}{\textit{Night}}\\
 \cline{2-7}
 & \textit{3m} & \textit{5m} & \textit{3m} & \textit{5m} & \textit{3m} & \textit{5m} \\\hline
\textit{Single} & 0.974 & 0.988 & 0.627 & 0.827 & 0.392 & 0.535 \\
\hline
\textit{SeqM} & 0.980 & 0.994 & 0.736 & 0.905  & 0.538 & 0.737 \\
\hline
\textit{Graph} & 0.990 & 0.996 & 0.728 & 0.912  & 0.581 & 0.737 \\
\hline
\textit{Topological} & 0.990 & 0.997 & 0.893 & 0.996 & 0.850 & 0.975 \\
\hline
\textit{MCL} & \textbf{0.998} & \textbf{0.999} & 0.928 & 0.993 & 0.844 & 0.981 \\
\hline
\textit{MCL (RTK)} & \textbf{0.998} & \textbf{0.999} & \textbf{0.953} & \textbf{1.000} & \textbf{0.925} & \textbf{0.992}\\
\hline
\end{tabular}
\end{center}
\vspace{-7mm}
\end{table}

\begin{table}[ht]
\caption{Area-under-curve (DenseVLAD)}
\vspace{-3mm}
\label{tab:aucDV}
\begin{center}
\begin{tabular}{|c||c|c||c|c||c|c|}
\hline
\multirow{2}{*}{Model} & \multicolumn{2}{c||}{\textit{Rain}} & \multicolumn{2}{c||}{\textit{Dusk}} &\multicolumn{2}{c|}{\textit{Night}}\\
 \cline{2-7}
 & \textit{3m} & \textit{5m} & \textit{3m} & \textit{5m} & \textit{3m} & \textit{5m} \\\hline
\textit{Single} & 0.958 & 0.987 & 0.717 & 0.824 & 0.506 & 0.651 \\
\hline
\textit{SeqM} & 0.980 & 0.993 & 0.835 & 0.965  & 0.714 & 0.870 \\
\hline
\textit{Graph} & 0.985 & 0.996 & 0.857 & 0.956  & 0.760 & 0.876 \\
\hline
\textit{Topological} & 0.990 & 0.997 & 0.905 & 0.997 & 0.881 & 0.983 \\
\hline
\textit{MCL} & 0.994 & \textbf{0.999} & 0.909 & 0.994 & 0.875 & 0.987 \\
\hline
\textit{MCL (RTK)} & \textbf{0.996} & \textbf{0.999} & \textbf{0.972} & \textbf{1.000} & \textbf{0.917} & \textbf{0.995} \\
\hline
\end{tabular}
\end{center}
\vspace{-5mm}
\end{table}

\subsection{Automatic scaling of localization latency}

We display the average number of steps to localize for each method in Table \ref{tab:stepstoloc} at the 99\% precision level. We can see that for our methods, lower levels of perceptual aliasing require fewer steps to localize on average which is not possible for the fixed length sequence matching approach.

\begin{table}[ht]
\caption{Avg steps to localize at 99\% precision (5m, 30 deg)}
\vspace{-3mm}
\label{tab:stepstoloc}
\begin{center}
\begin{tabular}{|c||c|c||c|c||c|c|}
\hline
\multirow{2}{*}{Model} & \multicolumn{2}{c||}{\textit{Rain}} & \multicolumn{2}{c||}{\textit{Dusk}} &\multicolumn{2}{c|}{\textit{Night}}\\
 \cline{2-7}
 & \textit{NV} & \textit{DV} & \textit{NV} & \textit{DV} & \textit{NV} & \textit{DV} \\\hline
\textit{Single} & 1.0 & 1.0 & 1.0 & 1.0 & 1.0 & 1.0 \\
\hline
\textit{SeqM} & 10.0 & 10.0 & 10.0 & 10.0 & 10.0 & 10.0 \\
\hline
\textit{Graph} & 5.3 & 4.6 & 12.2 & 9.4 & 10.7 & 11.5 \\
\hline
\textit{Topological} & 6.4 & 6.1 & 9.8 & 9.9 & 17.5 & 12.9 \\
\hline
\textit{MCL} & 5.5 & 5.6 & 13.9 & 12.3 & 18.2 & 16.1 \\
\hline
\textit{MCL (RTK)} & 5.6 & 5.6 & 8.9 & 7.4 & 17.4 & 13.8 \\
\hline
\end{tabular}
\end{center}
\vspace{-5mm}
\end{table}

\subsection{Computation Time}

Finally, we benchmark the computation time per iteration for each of our methods at 99\% precision. All methods were implemented in Python using the Numpy library, and all computations were vectorized as much as possible within a single trial. Benchmarks were run on a desktop with a Intel® Core™ i7-7700K CPU, 32Gb RAM running Ubuntu 20.04 and the results are shown in Table \ref{tab:computationtimes} for the 5m, 30 deg error tolerance, where reasonable recall performance was achieved across all methods. 

On a cost per iteration level, our proposed MCL algorithm is the most expensive. This is mostly attributed to the nearest neighbor search from \eqref{eq:sensorupdateapprox} as well as the particle motion update. In contrast, our topological localizer is comparable to both single image matching and the linear sequence matching approach due to the simple, parallel and vectorizable nature of the underlying computations. Finally, graph-based matching is a serial algorithm, resulting in more expensive iterations overall compared to our topological localizer. 

\begin{table}[ht]
\caption{Compute time per iteration (ms) at 99\% precision (5m, 30 deg)}
\vspace{-3mm}
\label{tab:computationtimes}
\begin{center}
\begin{tabular}{|c||c|c||c|c||c|c|}
\hline
\multirow{2}{*}{Model} & \multicolumn{2}{c||}{\textit{Rain}} & \multicolumn{2}{c||}{\textit{Dusk}} &\multicolumn{2}{c|}{\textit{Night}}\\
 \cline{2-7}
 & \textit{NV} & \textit{DV} & \textit{NV} & \textit{DV} & \textit{NV} & \textit{DV} \\\hline
\textit{Single} & 9 & 9 & 9 & 9 & 9 & 9 \\
\hline
\textit{SeqM} & \textbf{9} & \textbf{9} & \textbf{9} & \textbf{9} & \textbf{9} & \textbf{9} \\
\hline
\textit{Graph} & 35 & 39 & 15 & 21 & 11 & 15 \\
\hline
\textit{Topological} & \textbf{9} & \textbf{9} & \textbf{9} & \textbf{9} & \textbf{9} & \textbf{9} \\
\hline
\textit{MCL} & 60 & 61 & 48 & 51 & 46 & 47 \\
\hline
\textit{MCL (RTK)} & 58 & 60 & 66 & 70 & 47 & 54 \\
\hline
\end{tabular}
\end{center}
\vspace{-5mm}
\end{table}

\subsection{Overall performance} We have presented two methods; a method that yields localization performance which outperforms existing state-of-the-art methods for comparable computation effort as well as another method which yields vastly superior performance given additional computation time and odometry estimates.
\section{Discussion and Conclusion}

We presented two new online, probabilistic sequence-based VPR back-ends that provide the tight error tolerances suitable for initializing a subsequent pose refinement step within a hierarchical localization system while achieving state-of-the-art results across various levels of appearance change on real-world datasets. The work here prompts a number of interesting avenues for future research. The question of how the localization load is optimally shared between the coarse localizer and the subsequent pose refinement step remains open; investigating how to determine this balance with respect to key factors such as performance and computational cost will likely yield further advances. 

In addition, one potential future application area for our proposed methods is within the visual teach and repeat literature \cite{furgale10, KRAJNIK2017, dallosto20}, specifically around vision-only localization in GPS denied environments \cite{warren19}. Our method may allow vision-only localization possible across severe appearance change with a \textit{single} mapping run, as opposed to resorting to an experience-based system \cite{paton16}.





\bibliographystyle{IEEEtran}
\bibliography{references,references_ieeectrl}

\end{document}